# A Hybrid Genetic Algorithm–Large Language Model Framework for Structured Task Optimization

William Shum, Rachel Chan, Jonas Lin, Benny Feng, Patrick Lau

## Abstract

Large Language Models (LLMs) have demonstrated a remarkable capacity to generate coherent text and to reason through complex problems. However, they often encounter difficulties with **structured tasks** that must obey rigid constraints or optimize multiple objectives at once. In this context, we introduce **GA-LLM**, which is a hybrid framework combining **Genetic Algorithms (GAs)** and LLMs. This approach is designed to tackle complex reasoning and structured text generation tasks that operate under explicit constraints. In GA-LLM, each structured output (for example, a detailed plan or a report) is treated as an individual "gene." The system then applies evolutionary strategies — namely selection, crossover, and mutation — guided by an LLM in order to iteratively refine the solutions. Within this hybrid loop, the LLM contributes domain knowledge and creative variation when generating or modifying candidates, while the GA component conducts a global search and optimization process that enforces structural constraints and drives the population towards high-quality solutions. We demonstrate the effectiveness of this approach on several tasks, including multi-day itinerary planning, academic proposal outlining, and business report generation. In these cases, GA-LLM automatically produces well-structured outputs that meet the specified requirements (such as budget limits or mandatory section coverage). The framework's modular design is another key advantage, making it straightforward to adapt GA-LLM to new tasks by defining custom gene representations, prompt templates, and constraints. Furthermore, we discuss how GA-LLM leverages the complementary strengths of evolutionary algorithms and LLM-based reasoning. We find that this combination can achieve more reliable constraint satisfaction and better solution optimization than relying on LLM prompting alone. In summary, **GA-LLM** offers a generalizable strategy for improving LLM-driven reasoning under strict structural requirements.

## 1. Introduction

Recent advances in **Large Language Models (LLMs)** have given rise to powerful systems capable of generating human-like text and tackling complex problems using their learned knowledge. These models can produce detailed outputs such as plans, code, or explanations from natural language prompts. However, a purely prompt-driven approach still often struggles when it comes to generating **structured outputs under strict constraints**. For instance, if one asks an LLM to plan a multi-day itinerary with tight budget and scheduling constraints, or to draft a formal document with required sections and length limits, it can be challenging for the model to get everything perfect in one pass. In many cases, LLMs might violate hard constraints (for example, exceeding the budget or leaving out a required section) or settle for solutions that are suboptimal, since they lack an explicit global search mechanism for optimization. Techniques like *chain-of-thought prompting* [1] and *tree-of-thought search* [2] have been proposed to improve reasoning by letting the model explore multiple steps or branches. While these methods do help by externalizing some reasoning, they still ultimately rely on the LLM's internal logic. As a result, even such approaches can get stuck in local optima or end up producing only partial solutions.

In parallel to these developments in LLMs, **Genetic Algorithms (GAs)** and other evolutionary algorithms have long been used for search and optimization in complex problem spaces. GAs work by maintaining a **population** of candidate solutions and iteratively refining them via processes inspired by natural selection. Typically, this involves evaluating a **fitness** function for each candidate, selecting the fittest individuals, and generating new candidates through **crossover** (recombining parts of two solutions) and **mutation** (randomly modifying parts of a solution) [3]. This population-based approach has proven particularly effective at exploring large, **multi-modal search spaces** and escaping local optima that might trap simpler algorithms. However, traditional GAs require a well-defined encoding for solutions and a computable fitness function to evaluate them. In tasks involving natural language or other structured text, it is non-trivial to design an effective numerical fitness function. Moreover, a purely GA-based approach may generate a lot of *invalid or nonsensical solutions* if there isn't domain knowledge guiding the variations, especially when dealing with the richness of language.

To address the limitations of each approach alone, we propose **GA-LLM**, which marries the strengths of GAs and LLMs into a **hybrid optimization framework** for structured reasoning tasks. The key insight behind GA-LLM is to employ an LLM's generative and evaluative capabilities at each step of an evolutionary search process. In practice, GA-LLM represents each candidate solution in a structured format – for instance, an itinerary could be represented as a list of days with activities, or a report as an outline with sections. The LLM is invoked to **generate** an initial set of candidate solutions and also to perform **crossover and mutation** in a semantically informed way. This means that when new offspring solutions are created, the LLM helps ensure they remain coherent and adhere to the task format. The LLM is furthermore used to **evaluate** the candidates (for example, by scoring how well an itinerary meets user preferences or how complete a proposal draft is), thereby providing a flexible kind of fitness function even in cases where no simple numeric evaluation metric exists. Meanwhile, the GA component orchestrates the overall search: it selects the best candidates according to these LLM-based evaluations and enforces any hard constraints via filtering or penalization. Over successive generations, the population **converges towards higher-quality solutions** that satisfy all the structural constraints and optimize the desired criteria.

By integrating an LLM into the evolutionary loop, GA-LLM benefits from both the LLM's *knowledge and creativity* — which helps generate diverse, content-rich solutions and meaningful variations — and the GA's *robust search ability* — which systematically explores many possibilities and refines them. This combined approach is able to handle tasks where a single-pass LLM output might often fall short or would otherwise require extensive manual prompting to improve. We illustrate the capabilities of GA-LLM through several use cases. These include: a travel itinerary planner that must optimize for cost and scheduling constraints, an academic proposal generator that ensures all required sections are present and consistent, and a business report builder that respects a given template. In each case, GA-LLM produces structured outputs that adhere to the specified constraints (such as budget limits or mandated section headings) while simultaneously optimizing overall quality metrics that we provide to the LLM (for example, user satisfaction scores or coverage of required topics).

The contributions of this work are as follows: (1) We formulate **GA-LLM**, a general **hybrid framework** that combines evolutionary search with LLM-based generation and evaluation for complex structured tasks. (2) We implement a **modular architecture** with an abstract gene representation, which allows easy extension to new tasks by specifying domain-specific prompts and constraints. (3) Through example applications, we demonstrate that GA-LLM can automatically generate solutions to non-trivial reasoning problems (such as itineraries and outlines) that satisfy constraints and objectives more effectively than naive single-pass prompting. (4) We discuss how GA-LLM relates to and extends existing strategies for LLM-based reasoning, offering a novel avenue to leverage LLMs within optimization loops. Additionally, we have released an open-source implementation of GA-LLM to the community, which can be adapted and built upon for further research.

The remainder of this paper is organized as follows. **Section 2** reviews related work on improving LLM reasoning and surveys prior attempts at integrating evolutionary algorithms with LLMs. **Section 3** details the GA-LLM framework itself, including the problem formulation, system architecture, and the role of each component. **Section 4** presents example applications and qualitative results, illustrating how GA-LLM performs on representative structured tasks. **Section 5** offers concluding thoughts and outlines directions for future work. Finally, we note that the GA-LLM code is available as open source (see the Code and Resources section), which is provided to facilitate further research and development.

## 2. Related Work

**LLM Reasoning and Prompting Techniques:** As LLMs have grown more capable, researchers have explored various methods to induce better reasoning and stricter constraint adherence in their outputs. One straightforward approach is *prompt engineering*, where carefully designed instructions or structured examples are provided to the model. While this can sometimes yield improved results, it often still produces only a single-shot output with limited opportunity for adjustment.

The **chain-of-thought (CoT)** approach [1] demonstrated that prompting an LLM to generate intermediate reasoning steps can significantly improve its performance on arithmetic and logical problems by breaking them down into smaller pieces. Building on a similar idea, the **Tree-of-Thoughts (ToT)** framework [2] goes a step further by performing a search over possible multi-step reasoning paths. In ToT, the LLM proposes multiple solution steps at each stage, and an external search algorithm (e.g., a depth-first or breadth-first search) explores different combinations of these steps. Partial solutions are evaluated along the way to decide which branch to expand next. These methods show that introducing an external **search or iterative refinement** component can bolster an LLM's problem-solving abilities beyond what a single pass can do.

Our GA-LLM shares a similar spirit of externalizing the search process; however, instead of searching over sequential reasoning steps as CoT or ToT do, we search over **complete candidate solutions** in parallel using evolutionary strategies. There are also other iterative refinement techniques such as **self-reflection** or self-editing loops, where an LLM critiques and revises its own output iteratively [6]. For example, the *Self-Refine* method [6] lets an LLM provide feedback on its initial answer and then generate a revised output. This process can repeat multiple times to incrementally improve quality without any human intervention. GA-LLM's approach is different in that it maintains a **population of multiple solutions** simultaneously. This encourages exploring a diverse set of possibilities rather than repeatedly refining a single answer. Such a population-based approach can help the model avoid getting stuck in a limited solution space, because the algorithm can recombine parts of different promising outputs to discover new and potentially better solutions.

**Evolutionary Algorithms in NLP and LLM Contexts:** Evolutionary algorithms, including GAs, have also been applied to various problems in natural language processing and text generation, especially when the solution space is combinatorially large. Historically, some early attempts used GAs for generating text or for solving tasks like scheduling and planning described in natural language, though these often had to rely on very simplified language models or rigid templates due to limitations of the technology at the time. In recent years, interest has grown in directly combining **LLMs with evolutionary algorithms**. For example, Guo et al. (2024) [4] demonstrate that *connecting LLMs with evolutionary algorithms yields powerful prompt optimizers*. In their framework, candidate prompts are evolved using crossover and mutation operations on the text of the prompts, and those that lead to better task accuracy are selected for the next generation. Similarly, **PhaseEvo** by Cui et al. (2024) [5] employs a two-phase genetic optimization of prompts: first a broad exploration phase with larger mutations, followed by a fine-tuning phase with more focused edits. Both of these works treat the LLM's input prompt itself as the subject of evolutionary optimization.

In contrast, GA-LLM treats the LLM's *output content* (for a given task prompt) as what is being evolved. In our framework, the LLM remains central to generating and refining content, but the **object of optimization is the content itself** (e.g., the text of an itinerary or proposal draft), guided by a fitness measure of quality.

Another line of related research integrates evolutionary search to improve LLM performance on problem-solving tasks. Jimeno-Yepes and Barnard (2025) [7], for instance, present an approach where LLMs are used within an evolutionary algorithm to generate candidate solutions (particularly for coding or reasoning tasks) and then a fast evaluation mechanism filters and scores these solutions to handle large populations efficiently. Their work highlights how evolutionary algorithms can effectively traverse the huge search space of possible LLM outputs, and it shows that LLMs can generate more meaningful candidates than random initialization. GA-LLM subscribes to this philosophy by ensuring that every candidate in the population is either *LLM-derived or LLM-refined*. This means candidates are likely to be syntactically correct and semantically plausible, which helps accelerate convergence compared to blind random mutations.

Additionally, there have been ideas borrowed from large-scale distributed search in contexts like code generation. For example, DeepMind's **AlphaCode** system generated a vast number of program samples and then filtered and clustered them to identify correct solutions for coding challenges [8]. While AlphaCode did not use genetic recombination or mutation (so it isn't a genetic algorithm per se), it still illustrates the power of a **generate-and-select** paradigm in conjunction with a learned model. In a sense, GA-LLM can be seen as a principled *generate-select-refine* paradigm, which introduces recombination and iterative improvement into that loop.

In summary, GA-LLM builds upon these prior ideas of enhancing LLM outputs via external search and optimization. Our work is distinguished by focusing specifically on **structured text generation tasks** (moving beyond just prompt optimization) and by providing a general-purpose, extensible framework in which any structured task can be plugged in simply by defining how to represent solutions and how to interact with the LLM. To the best of our knowledge, GA-LLM is among the first frameworks to explicitly combine genetic algorithms with LLM-based content generation for complex, constrained writing and planning tasks. Next, we describe the architecture and methodology of GA-LLM in detail.

## 3. GA-LLM Framework and Methodology

GA-LLM is a **modular hybrid framework** that brings together the iterative optimization process of a genetic algorithm with the generative and evaluative power of a large language model. In this section, we describe in detail how a structured task is formulated within GA-LLM and how the LLM and GA components interact throughout the process. Figure 1 provides a high-level flowchart of the GA-LLM process, illustrating the progression from initial population creation to evolutionary refinement.

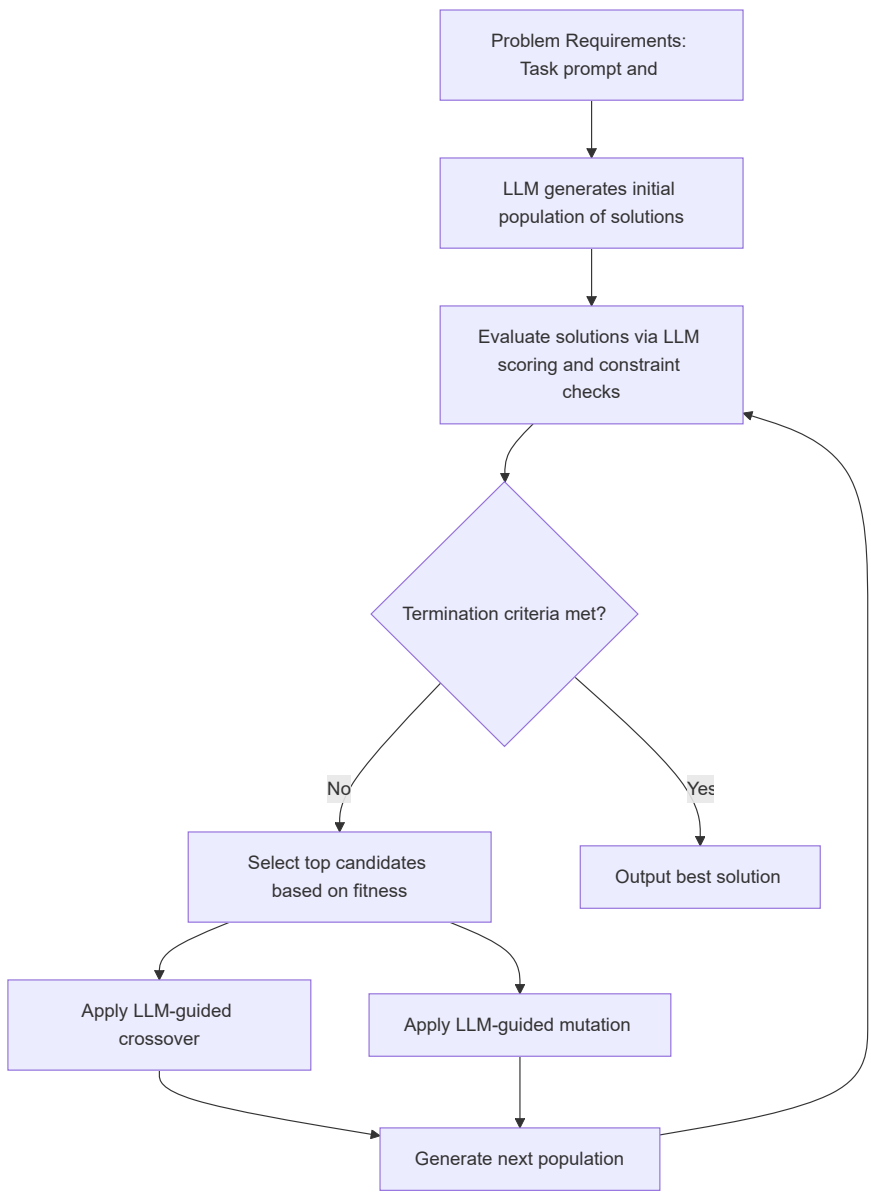

*Figure 1: Flowchart illustrating the GA-LLM hybrid optimization process.* The algorithm begins by using an LLM to generate an **initial population** of candidate solutions that satisfy the given task prompt and basic requirements. It then enters an evolutionary loop. In each generation, every candidate is **evaluated** (either by the LLM or by auxiliary functions) to assess its quality or fitness, and any defined **constraints** are checked. If the stopping condition has not yet been met (for instance, if the maximum number of generations is not reached or the solution quality is still below a desired threshold), the algorithm **selects** the top-performing candidates. Those selected candidates serve as parents to **produce new candidates** via genetic operations: parts of two solutions can be combined (**crossover**), and parts of a solution can be modified (**mutation**). These operations are guided by the LLM to help ensure that the resulting offspring solutions are valid and creative. The newly generated candidates then form the next generation of the population. This generation is subsequently evaluated, and the loop repeats as described. When the termination criteria are finally satisfied, the best solution found in the population is output as the result.

### 3.1 Task Representation and Gene Abstraction

One cornerstone of GA-LLM is the use of a **task-aware encoding** for solutions. In practice, we define a custom **gene representation** for each target task, which captures the structured nature of solutions in that domain. Formally, one can think of a task's solution space as the set of all valid structured outputs for that task (for example, all possible travel itineraries of a certain length, or all possible outlines for a research proposal). In GA-LLM, an individual candidate solution is represented as an instance of a domain-specific class (for example, a class `TravelGene` or `ProposalGene` for those tasks) that inherits from an abstract base class `BaseGene`. This gene object contains fields or components corresponding to parts of the structured output (e.g., a list of day plans for an itinerary, or the sections of a report in an outline).

Each gene class must implement certain methods to interface properly with both the LLM and the GA. In particular, methods like `to_text()`, `parse_from_text()`, `crossover()`, and `mutate()` are required. The `to_text()` method serializes the gene (i.e., the structured data) into a textual format that can be fed to the LLM—for example, for evaluation or as part of a prompt for further generation. Conversely, `parse_from_text()` interprets text generated by the LLM and converts it back into the structured form of the gene. These two functions handle the translation between the **semantic structured representation** of a solution and the **textual natural language** that the LLM operates on. By explicitly parsing LLM outputs in this way, GA-LLM can enforce that only well-structured, valid solutions become gene instances in the population. If the LLM happens to output ill-formed text that doesn't conform to the expected format, the parser will detect this, which leads to that candidate being rejected or repaired. This mechanism keeps the evolutionary search on track by ensuring that the population consists mostly of valid individuals.

The gene class also implements the standard **genetic operators**. For example, the `crossover(parent_gene)` method defines how to produce an offspring gene by combining this gene with another parent, and `mutate()` defines how to randomly alter parts of the solution. These operations can be implemented in various ways depending on the nature of the task. A straightforward approach is to perform **direct structural crossover or mutation** on the genes: for instance, one could swap some sub-components between two parent solutions, or randomly change a component of a single solution. However, in many cases we leverage the LLM to perform these genetic operations more *intelligently*. For example, a crossover could be executed by prompting the LLM with two parent solutions along with an instruction to "combine the best aspects of these into a new solution." The LLM would then produce a merged output, which we parse into a new gene. Likewise, a mutation can be guided by the LLM by giving it a single solution and an instruction like "modify this solution slightly to meet X, or introduce some variation in Y." Using the LLM for these variation operators helps ensure that the offspring solutions remain coherent and high-quality. The LLM can effectively rephrase or adjust content in a natural way, rather than producing jarring random changes as a naive mutation might.

This approach contrasts with traditional GA mutations that randomly perturb an encoded solution without understanding. Here, the mutations are more **semantic** and often informed by the task context (for example, changing an activity in a travel plan to a different one that serves a similar role, or rewriting a paragraph in a proposal to shift its focus slightly). The framework's **prompt templating system** supports this advanced use of the LLM by allowing users to define predefined templates for generation, crossover, and mutation prompts. These templates include placeholders where current candidate content is inserted as needed. Users can customize these templates to control how the LLM should produce new variants, giving a degree of control over the style and nature of the generated solutions.

## 3.2 Fitness Evaluation via LLM and Constraints

Defining a robust **fitness function** is crucial for guiding the genetic algorithm's search. GA-LLM provides a flexible **LLM-based scoring mechanism** [7] that enables evaluation of complex criteria which would be difficult to encode mathematically. For each candidate solution (gene), the system constructs a **scoring prompt** that is fed to the LLM, asking it to assess the solution along relevant dimensions. For example, in the travel itinerary task, the prompt might be something like: *"Here is a 4-day travel itinerary with cost details. Please rate this itinerary on a scale of 1 to 10 for overall quality, considering criteria such as logical scheduling, variety of activities, and adherence to the budget. Provide a brief explanation for the score."* The LLM's output can then be parsed to extract a numerical score. (Optionally, the accompanying explanation can be used to verify that the evaluation was reasonable.) In an academic proposal task, similarly, the LLM might be asked to judge how well the draft covers all required sections or how innovative the proposed idea is. In effect, this approach uses the LLM as a **domain-aware fitness estimator**, leveraging the model's knowledge to judge qualitative aspects of a solution (like coherence, relevance, or creativity) that would be very hard to capture with a hand-crafted numeric formula. It also allows what might be multiple objectives to be combined into a single overall rating guided by natural language criteria. We note that prompt design for evaluation is important — the prompt should be crafted in a way that reduces bias and variability in the LLM's scoring. For example, providing a consistent rubric or examples of how to evaluate can help get more stable and fair assessments from the model.

In addition to the LLM-based soft evaluation, GA-LLM incorporates a **constraint validator** module to handle any *hard constraints* that solutions must satisfy. These constraints can include structural requirements (for example, an itinerary might be required to have exactly 4 days, each with a certain format of activities; or a business report might be required to include an Executive Summary section) as well as domain-specific limits (for instance, the total cost must not exceed $5000, or a proposal must not be longer than 2 pages). The framework allows users to plug in custom constraint-checking code for each task, either by implementing a specified interface or by providing validation functions. During the evaluation phase, each candidate solution is run through this validator. If any hard constraint is violated, the system either assigns that candidate a very low fitness score (making it unlikely to be selected), or simply filters the candidate out from the population immediately. By integrating this step, GA-LLM ensures **feasibility** of solutions: the evolutionary search is effectively constrained to focus on the subspace of solutions that meet all the mandatory conditions. This is a critical advantage over naive LLM generation, which might produce an output that looks good superficially but is unusable because it violates a key requirement. In GA-LLM, any such invalid output is promptly caught and either discarded or repaired. (For example, if the only issue is a minor formatting error, the system could attempt an automatic correction or even ask the LLM to fix the format.)

## 3.3 Evolutionary Search Process

Given the representation, variation operators, and evaluation scheme described above, GA-LLM runs a classic evolutionary loop with these components. The **HybridEvolutionEngine** is the core engine that manages this process. First, an initial population of $N$ solutions is generated. Rather than starting from scratch with completely random solutions (which, in a complex structured space, would likely yield many useless individuals), we leverage the LLM to create a diverse initial population. For example, we can prompt the LLM $N$ times (introducing some randomness or varying the phrasing each time) to "propose a solution" for the task. In the travel planning scenario, we might ask the LLM something like: *"Generate a 4-day trip itinerary for Shanghai under a budget of ¥5000"*, multiple times. By varying the prompt wording or using a higher randomness setting (temperature), we obtain a variety of different itineraries. These initial textual outputs are then parsed into gene instances. By seeding the population with LLM-generated individuals, we ensure a good baseline of **feasible, reasonably high-quality solutions** that already meet basic requirements (since we included the constraints in the prompt itself).

Once the initial population is ready, the evolutionary loop proceeds through repeated generations. In each generation, every individual in the population is **evaluated** by applying the constraint checker and by querying the LLM for a fitness score as described earlier. This yields a fitness value for each candidate (and some candidates might be disqualified outright if they fail hard constraints). GA-LLM supports **parallelized evaluation** using thread pools, meaning that multiple LLM calls for scoring can happen concurrently—this speeds up the evaluation phase, which is especially helpful if the population size $N$ is large or if the LLM calls are slow.

After evaluation, a **selection** mechanism decides which individuals will serve as parents for producing the next generation. In our implementation, we typically use an **elitist selection** strategy combined with either tournament selection or rank-based selection. That means a fixed small number of top-scoring individuals (the "elite" survivors) are carried over unchanged, and additional parent individuals are probabilistically chosen from the population with probabilities weighted by fitness. (The selection strategy and parameters such as population size, number of elites, mutation probability, etc., are configurable in the framework's settings.) Preserving a few elite solutions ensures that the best solutions found so far are not lost, while the probabilistic selection injects some diversity and gives a chance for other good solutions to contribute to the next generation.

Next, the **crossover** operation is applied to the selected parents to produce new offspring. Because our crossover procedure may rely on calling the LLM (which might produce only one offspring per call), we typically iterate over parent pairs and generate one or two offspring for each pair via the `crossover` method. For example, consider two parent itinerary genes, A and B. We could call the LLM with a prompt that includes both A's and B's itineraries and instructs it: *"Combine elements of these two itineraries into a new 4-day itinerary, ensuring it stays within the budget."* The LLM's output in response to this prompt can be parsed into a new itinerary gene, C. This new gene C might incorporate, say, some days from A and some from B, or a mix of attractions from both, phrased in a coherent way.

We also perform **mutation** on some individuals in each generation, typically on the newly created offspring (and occasionally we might even mutate a surviving elite to explore variations). For instance, a mutation prompt for an itinerary might be: *"Here is a 4-day itinerary. Modify it by changing one activity or adjusting the schedule, while keeping it coherent."* The LLM then returns a slightly altered itinerary — for example, it might replace a planned museum visit with a visit to a park on Day 2, or swap the order of two activities on a given day. Such mutations introduce new diversity into the population and help the algorithm explore parts of the solution space that were not covered by the current parents. This diversity is essential for avoiding premature convergence of the population.

After performing crossover and mutation, we obtain a set of new candidate solutions that form the **next generation**. We generally design the process so that the population size remains constant from generation to generation. In other words, if we started with $N$ individuals, then each generation we produce $N$ offspring (replacing the entire population). The framework could also support a steady-state approach (where only some individuals are replaced each generation), but in our current implementation we use a simple generational model for simplicity. This new generation then goes through the same cycle of evaluation, selection, and variation as described above.

This loop of evaluate-select-reproduce continues until a specified **termination condition** is met. Common termination criteria include reaching a maximum number of generations $G_{\max}$, finding a solution that exceeds a desired fitness threshold (for example, an itinerary that receives a perfect 10/10 score, or a proposal draft that meets all criteria perfectly), or detecting stagnation in the population (such as the top fitness not improving for many generations). The framework allows the user to specify these stopping conditions as part of the configuration. In many automated scenarios, a fixed number of generations is chosen for simplicity. When termination conditions are satisfied, GA-LLM outputs the best solution found during the run (or possibly a set of top solutions, especially if one desires multiple diverse answers).

Throughout the evolutionary run, GA-LLM utilizes a **multi-level logging** system. Each run can produce a log that records key events and metrics for each generation — for example, the highest fitness value in the generation, a summary of any constraint violations encountered, and perhaps snippets or descriptions of the top solutions. This logging is useful for analyzing how the optimization process is proceeding and for debugging the behavior of prompts or genetic operators. By examining the logs, one can see details such as whether the LLM's evaluations are consistent over time, or whether a crossover operation is blending solutions in the expected way. The logging verbosity is configurable, allowing users to balance the level of detail recorded with the performance overhead of logging.

## Complexity and Performance Considerations

Using an LLM within a GA loop does introduce additional computational overhead, as LLM calls are typically expensive (in terms of time and possibly cost, if using a paid API). GA-LLM mitigates this overhead via parallelization: as noted, scoring and candidate generation can be parallelized since each candidate's evaluation or creation is independent of the others. If one is using a large remote LLM, a practical approach is to limit the population sizes and the number of generations to what is feasible within an acceptable runtime. Interestingly, we have observed that because the LLM provides a much smarter search heuristic than random guessing, even a relatively small population (on the order of tens of individuals) and a modest number of generations can be enough to significantly improve solution quality.

The LLM's ability to make large leaps in solution space — for instance, rewriting a poor solution into a much better one in a single mutation step — means that the GA does not have to rely solely on tiny incremental changes. This often accelerates convergence compared to what a traditional GA would require on a similar problem. In fact, in Section 4 we provide an example illustrating GA-LLM's convergence behavior on a planning task, showing how quickly it can reach a high-quality solution.

Another practical consideration is the **stochasticity** introduced by the LLM. Calls to an LLM can produce different results from one run to the next, especially if we allow non-zero randomness in generation (which is often useful for exploration). This inherent randomness can be beneficial because it injects diversity into the search (similar to how random mutations work). However, it also means there can be variance in fitness evaluations — the same solution might receive slightly different scores in different evaluations if the LLM's scoring isn't completely deterministic. To counteract this, one can prompt the LLM in a way that encourages determinism for the evaluation step (for example, using a fixed low temperature or phrasing the prompt to encourage a consistent evaluation rubric). Another approach is to evaluate each candidate multiple times and average the scores if very high precision is needed for selection. In our current framework design, we assume a single-pass scoring for each candidate to keep things efficient, but advanced users could certainly extend this to more robust evaluation strategies if needed.

## 3.4 Illustrative Example

**Task setup:** For concreteness, consider the task of **Travel Itinerary Planning**, which we include as an example demo with the GA-LLM framework. The goal here is to automatically generate a detailed itinerary for a trip (say, 4 days in a particular destination) that fits certain user-specified constraints (e.g., staying under a budget limit and covering a set duration) and ideally maximizes the traveler's experience (ensuring a mix of interesting places without an overly exhausting schedule). The structured solution format for this task can be defined as a list of days, where each day has a date and a list of activities (including time, location, and a brief description), along with summary information like the total cost and the chosen hotel or accommodations.

**Process:** We implemented a `TravelGene` class to represent an itinerary. Each itinerary gene contains fields for each day's schedule, the total cost, and so on. The **initial population** of itineraries for a query like "4-day trip to Destination X under budget Y" is generated by prompting the LLM to create complete itineraries. These initial solutions vary widely: some might allocate more of the budget to hotels while others spend more on attractions; some might put famous tourist spots all in the beginning, whereas others spread them out or leave them for later days. All of the initial itineraries are at least plausible itineraries, but naturally some are better than others in terms of coherence, efficient use of time, or cost-effectiveness.

**Results:** Many of the initial LLM-generated plans had room for improvement. For instance, some initial itineraries violated the budget or had an uneven distribution of activities (perhaps cramming too many activities into one day and too few in another). After a few generations, however, the GA-LLM process produced itineraries that were both **feasible** and **high-quality**. Our evaluation prompt in this case asked the LLM to score each itinerary based on factors like how well it covers diverse attractions, whether it stays under the budget, and if the daily schedule is logical. Meanwhile, our constraint checker automatically enforced the hard rules: e.g., total cost must not exceed the budget Y, and the itinerary must have exactly 4 days with reasonable start and end times for each day. Any itinerary that went over budget was either penalized in its score or removed entirely. The GA then tended to select itineraries that covered many key attractions while remaining under budget for breeding. For example, in one generation the algorithm might identify one itinerary (call it A) that had excellent content but was slightly over the budget, and another itinerary (B) that was under budget but missed some popular attractions. Through crossover, GA-LLM could produce a new itinerary (call it C) that takes some of the high-value activities from A but uses the cheaper accommodations or transportation choices from B to reduce costs, effectively combining their strengths. The LLM plays a crucial role in this step by merging details smoothly — for instance, if a day's schedule from A is adopted, the LLM adjusts references to ensure consistency with B's hotel or travel plans where necessary. Then a mutation might further refine itinerary C by, say, shortening an overly packed day or replacing an expensive activity with a free alternative, thereby further optimizing cost while retaining overall quality.

Over a few generations, we observed the population's best fitness score steadily improving. Eventually, GA-LLM returned an itinerary that met the budget and offered a well-rounded selection of activities. A simplified excerpt of one such resulting itinerary is shown below:

```
Travel Itinerary:
- **Days**: 4-day program
- **Hotel**: Shanghai Family Hotel
- **Total Cost**: ¥4890.00

**Detailed Schedule:**
Day 1: 2024-07-01
- 09:00 - Visit **Shanghai Museum** (historical art and artifacts)
- 11:30 - Lunch at **Yuyuan Old Street** (local street food)
- ... _(more activities)_

Day 2: 2024-07-02
- 10:00 - Explore **The Bund** (waterfront promenade)
- ...

... (Days 3 and 4 with activities) ...
```

*Example output (excerpt) from GA-LLM for a 4-day Shanghai itinerary.* In this sample output generated by our framework, the itinerary is well-structured (each day is clearly delineated with its activities and times) and it satisfies the budget constraint (the total cost ¥4890 is under the hypothetical ¥5000 budget). Notably, this itinerary was produced **automatically** by GA-LLM starting from an initial set of rough LLM-generated plans and then evolving them through the GA process. Many of the initial candidate itineraries were either over budget or poorly organized in some way. GA-LLM was able to refine those initial attempts via crossovers and mutations—guided by feedback from the LLM—into a high-quality plan. The final itinerary is something that an expert travel planner might have created, which shows that by combining the LLM's domain knowledge with the GA's optimization, the system can yield solutions that are both creative and feasible.

While the example above focuses on travel planning, the same GA-LLM framework can be applied to other structured tasks with appropriate customization. For instance, consider an **academic proposal writing** task. In that case, a gene could represent a proposal draft with sections such as *Introduction, Related Work, Methodology, Experiments, Conclusion*, etc. A crossover in this context might involve merging content from two proposal drafts—perhaps one draft has a stronger introduction while another has a more thorough methodology section, so the crossover produces a new draft that combines those strengths. A mutation might prompt the LLM to expand on a certain section or to rephrase a paragraph to improve clarity or add a missing detail. The fitness evaluation for proposals could be based on how well the draft covers all required points and the overall clarity or persuasiveness of the writing (as judged by an LLM or some heuristic), and the constraints might enforce things like page limits or the presence of all required sections.

Similarly, for a **business report generation** task, one could define a gene structure that includes key sections of the report (e.g., an Executive Summary, a Data Analysis section, Conclusions, etc.). GA-LLM could then optimize the content for completeness and coherence, while the constraints ensure that all mandatory sections are present in the final report. In preliminary experiments, we found that a GA-LLM based approach was able to satisfy structural requirements (like a company's formatting or section checklist) more reliably than a single-pass LLM solution, confirming the pattern that GA-LLM tends to outperform naive LLM generation when strict requirements must be met.

## 3.5 Comparison to Baseline Approaches

It is useful to contrast the GA-LLM approach with more conventional methods of using LLMs for structured tasks:

| Aspect | Direct LLM Prompting (single-shot or few-shot) | Iterative Refinement by LLM Alone | GA-LLM (Ours) |
|---|---|---|---|
| Solution Diversity | Limited – usually only one or a few outputs per prompt, making it hard to cover a broad range of possible solutions. | Limited – since it refines only a single solution path at a time, it may fail to explore alternative solutions. | High – it maintains a population of solutions in parallel, thus exploring multiple diverse candidates at once. |
| Constraint Satisfaction | Partially implicit – constraints are only given in the prompt and the LLM may ignore or forget some of them. | Implicit – constraints can be reinforced through iterative feedback to the model, but there's no guarantee they will be satisfied if the model gets stuck in a problematic solution. | Explicit – hard constraints are checked programmatically. Any invalid solutions are either fixed or removed each generation, providing a strong guarantee that the final output will meet all constraints. |
| Optimization of Objectives | Weak – a single-pass LLM output is often suboptimal, and there is no built-in mechanism to iteratively improve it except manually re-prompting. | Moderate – the LLM can be instructed to refine its answer in an iterative loop, but the improvement relies on the model's internal heuristics and it might even oscillate or worsen at times. | Strong – uses a defined fitness function to systematically improve the population over generations. Crossover allows combining good parts from different solutions, enabling hybrid solutions beyond what refining a single solution could achieve. |
| Use of Compute | Low per attempt – typically just one or a few forward passes of the model, which is computationally cheap for a single run. However, getting a good solution might require many tries with different prompts (manual re-prompting). | Moderate – requires multiple LLM calls in sequence on the same solution (for iterative refinement), which can be fairly expensive and time-consuming, though focusing on one solution at a time. | Higher – involves multiple LLM calls for many candidates in each generation, so it can be more computationally expensive overall. However, many of these calls can be executed in parallel, and the approach often finds high-quality solutions in fewer iterations than what brute-force repeated prompting would require. |
| Human Intervention | High – typically requires the user to adjust the prompt or try different prompts repeatedly if the output is not satisfactory. | Moderate – the user sets up the initial answer and an iterative refinement process. Some oversight may be needed to monitor results or adjust the feedback prompts if the refinement stalls or goes off track. | Low – after the user defines the task, prompts, and fitness criteria, the GA-LLM process runs autonomously. It doesn't require intervention as it automatically evolves and tunes the outputs. |

*Table 1: Qualitative comparison between GA-LLM and other strategies for constrained structured generation.* Direct prompting of an LLM may not fulfill all constraints or find the optimal solution without numerous manual retries. Iterative self-refinement loops allow the model to attempt improvements on its own, but they lack the robust exploration and the ability to recombine partial solutions that a GA provides. In contrast, GA-LLM strikes a balance between exploration and exploitation by evolving a population of candidates and explicitly enforcing constraints at each step. This comes at the cost of increased computational overhead, but that overhead is partly offset by parallelization and the guided search efficiency provided by the LLM's involvement.

## 4. Experiments and Case Studies

In this section, we present several illustrative applications of GA-LLM across different structured tasks. Instead of trying to benchmark on a single numeric metric (which is challenging for the open-ended tasks we consider), we focus on qualitative outcomes and discuss how GA-LLM improves solution quality over successive generations. All experiments were conducted using our GA-LLM implementation with a pre-trained large language model accessed via API (for example, GPT-4 or a model of similar capability, though our approach is generally model-agnostic). We crafted specific prompt templates for each task's generation, crossover, mutation, and evaluation steps as described earlier. The evolutionary parameters (such as population size and number of generations) were chosen empirically to balance solution quality with computational cost. In practice, we typically used population sizes on the order of 10–20 and ran the evolutionary loop for about 5–10 generations for the examples below, which we found sufficient to reach satisfying solutions.

### 4.1 Travel Itinerary Planning

**Task setup:** Plan a 4-day travel itinerary for Shanghai, China, under a budget of ¥5000 (approximately $700). The itinerary should include activities, accommodations, and an estimated total cost. The user desires a mix of cultural, historical, and leisure activities. The output needs to list each day's schedule with times and locations, as well as a summary of the total cost.

**Process:** We developed a `TravelGene` class to encode itineraries. The LLM generation prompt was crafted to request a 4-day itinerary in a structured JSON-like format, including cost estimates for each item. The fitness function combined multiple aspects: we asked the LLM to provide a score reflecting how enjoyable and well-balanced each itinerary was, and we also included a penalty in the score for any budget overflow. Hard constraints were set so that any itinerary exceeding ¥5000 or having fewer or more than 4 days would be considered invalid. GA-LLM was run with a population of 10 itineraries for 5 generations in this experiment.

**Results:** The initial itineraries generated by the LLM covered a range of styles – for example, some focused heavily on museums and temples, while others emphasized shopping and food. The total cost of the initial plans varied from very cheap (¥3000) to over the budget (¥6000). Many of the first-generation plans violated the budget constraint or had an uneven distribution of activities (e.g., too many attractions crammed into one day and too few on another). After a few generations, GA-LLM evolved itineraries that were both **feasible** (meeting all constraints) and **high-quality**. The best solution (an excerpt of which was shown earlier in Section 3.4) ended up with a total cost of ¥4890, staying under the ¥5000 budget, and it included a balanced mix of attractions. For instance, it combined historical sites like the Shanghai Museum and Yu Garden, cultural experiences such as local markets, and modern sights like the Shanghai Tower observation deck. The daily schedules in the final itinerary were well-paced, typically starting around 9–10 AM and ending by evening, with minimal travel time between consecutive activities. Notably, the LLM seemed to naturally cluster geographically nearby sites on the same day as it refined the itineraries – a detail that emerged as good elements from different candidates were combined.

Through the evolutionary process, we observed that **crossover operations** would often take an economical choice from one itinerary and insert it into another itinerary that had higher costs. For example, one crossover replaced an expensive hotel in one itinerary with a more affordable hotel from another itinerary, instantly reducing the total cost while preserving most of the activities of the high-cost plan. Mutations typically adjusted one part of the itinerary: in one case, a mutation swapped out a pricey river cruise for a free stroll along the Bund waterfront, and the LLM described this new activity in an appealing way. This change reduced the cost and was rewarded by the fitness function. The LLM's involvement was crucial to ensure that after such modifications, the itineraries still read coherently — for example, when the river cruise was replaced with a walk, the LLM made sure the narrative flowed and references were consistent (e.g., if the cruise included a dinner, after replacing it with a walk, that dinner reference was removed or changed appropriately). By generation 5, nearly all itineraries in the population were under budget and included nearly all of the "must-see" locations that had been scattered across different initial candidates. This indicates a successful **convergence**: advantageous traits (such as cost-effective planning and inclusion of popular attractions) had propagated throughout the population.

### 4.2 Academic Proposal Drafting

**Task setup:** Generate a structured outline for a research project proposal in the field of artificial intelligence, including sections for Introduction, Related Work, Methodology, Experiments, and Conclusion, with a coherent narrative and no more than 2 pages of text. The proposal should present a novel idea and clearly situate it in the context of prior work.

**Process:** We defined a `ProposalGene` class in which the gene structure explicitly represents the proposal with the named sections as components. We prompted the LLM to generate initial proposal drafts given a short description of the research idea (for example, *"an AI system for healthcare diagnostics"* as a topic prompt). The fitness evaluation relied on the LLM to judge each draft's clarity and completeness (checking, among other things, whether all required sections were present and well-developed), as well as the novelty of the proposed idea. We also implemented a simple hard constraint requiring that all five section headings (Introduction, Related Work, etc.) appear at least once in the draft. The population size for this experiment was set to 8 proposals, and we ran GA-LLM for 5 generations.

**Results:** The initial proposals produced by the LLM often had some shortcomings. For example, a few drafts did not clearly separate the Related Work section (it was sometimes merged into the Introduction), and some had a very superficial Experiments section. As GA-LLM evolved the population, we saw a cross-pollination of content between drafts. One proposal had a particularly strong Methodology section with detailed algorithms and evaluation metrics, which another draft was lacking; through crossover, that detailed methodology was inserted into the other proposal's structure. Conversely, a different proposal draft had a very well-written Introduction that framed the problem nicely, and that introduction got transferred into some drafts that initially had weaker openings. Mutations guided by the LLM were used to flesh out missing pieces — for instance, if a proposal lacked detail in the Experiments section, we would prompt the LLM (via a mutation operation) to elaborate on an experiment design for that proposal, effectively expanding that section in the next generation.

By the final generation, the top proposal outline was substantially improved compared to any single initial draft. It cleanly separated the five required sections, and each section contained substantive content. The Introduction clearly stated the problem and the contributions of the proposal; the Related Work section, although kept at a summary level for our experiment, did mention relevant areas; and the Methodology section was detailed and concrete. The LLM's evaluation scores for this final draft were consistently high — roughly equivalent to a 9 out of 10 or an "excellent" rating on our criteria for clarity and completeness (whereas the initial drafts had averaged around 6 out of 10). Moreover, by this stage all candidate proposals in the population satisfied the structural constraint of having all the required sections.

This experiment highlights GA-LLM's strength in enforcing **structural consistency** and improving **content coverage**. By exploring multiple draft solutions in parallel and exchanging their best parts, the GA-LLM process was able to produce a proposal outline that arguably none of the initial single-pass LLM outputs would have achieved on their own. This hints at the potential of GA-LLM for tasks like automated report generation or assisting human writers by intelligently merging and refining drafts.

### 4.3 Further Applications

We briefly note that GA-LLM can be and has been applied to other domains as well. For example, we have experimented with **business report generation**, where the goal is to ensure that all key elements of a report (such as an Executive Summary, Financial Analysis, Conclusions, etc.) are present and well-polished. We also explored **structured workflow or code generation** tasks, where the output might be a sequence of steps or pseudocode that must satisfy logical constraints. In these initial explorations, the results have been promising. For instance, a GA-LLM-driven business report generator was able to follow a company's prescribed formatting rules and include all required sections more reliably than a single-pass LLM, which often missed some items from the checklist. We have not yet conducted a detailed quantitative evaluation in these domains, as each would require task-specific metrics to fully assess. However, the qualitative pattern we observe across these case studies is consistent: GA-LLM tends to outperform a naive LLM approach in terms of meeting hard requirements and optimizing for various soft objectives. This is largely thanks to the evolutionary search guiding the LLM's creative generation process.

## 5. Conclusion and Future Work

We have presented **GA-LLM**, a novel hybrid framework that integrates genetic algorithms with large language models to tackle complex structured generation and reasoning tasks under constraints. GA-LLM leverages the **complementary strengths** of these two paradigms: the former's robust global search and optimization capabilities and the latter's rich world knowledge and language generation skills. Through examples like itinerary planning and document drafting, we demonstrated that GA-LLM can automatically produce high-quality solutions for tasks that satisfy structural constraints and optimized criteria more effectively than using an LLM alone. The framework's modular design allows it to be adapted to a wide range of tasks – users only need to plug in a new gene definition, craft appropriate prompts, and implement any task-specific validators to apply evolutionary optimization to their structured problem.

This work opens up several avenues for future research. First, a more **formal evaluation** on benchmark tasks would be valuable to quantify GA-LLM's improvements. For instance, one could measure the success rate of constraint satisfaction or quality metrics (perhaps via human evaluation) for GA-LLM vs. baseline methods across tasks. Another direction is exploring **improvements to the evolutionary strategy**: the current framework could incorporate other metaheuristics or selection schemes (e.g., multi-objective optimization if we care about multiple conflicting goals explicitly, or novelty search to maintain diverse solutions). The trade-off between LLM call cost and solution quality is also worth exploring – techniques to reduce the number of model queries, such as occasionally using cheaper surrogate models for partial evaluations or adopting a cleverer scheduling of evaluations, could enhance practicality. On the LLM side, as models improve and become more consistent, the reliability of LLM-based fitness evaluation will improve; one could also fine-tune an LLM to act as a more accurate evaluator for specific domains, thereby providing better guidance to the GA.

Another interesting future path is applying GA-LLM in interactive settings. For example, a human user could be kept "in the loop," reviewing top candidates each generation or injecting preferences (a form of human-influenced fitness function). This could turn GA-LLM into a powerful co-creative tool, where an AI and human collaboratively evolve solutions (be it a design, a piece of writing, or a plan). The general paradigm of *LLM-guided search* that GA-LLM exemplifies might extend beyond genetic algorithms to other evolutionary strategies or even brute-force search augmented by LLMs.

In conclusion, GA-LLM demonstrates a promising approach to **LLM-based problem solving**: by embedding an LLM's reasoning and generative capabilities within an evolutionary optimization framework, we can tackle tasks requiring both **creativity** and **rigorous constraint satisfaction**. We hope this work inspires further exploration of hybrid systems that unite model-based intelligence with algorithmic search to push the boundaries of automated reasoning and structured generation.

## Code and Resources

The GA-LLM framework has been released as an open-source project to encourage adoption and further development. The source code, along with demonstration tasks and documentation, is available at the **GA-LLM GitHub repository**: https://github.com/step2-technology/ga-llm. This repository contains the core library (which implements the genetic algorithm engine, base classes, and utility functions) as well as example scripts, including the travel itinerary demo described in this paper. We encourage users to refer to the README for instructions on defining new tasks, and we welcome contributions from the community in the form of improvements or new use-cases. The project is released under the MIT License (© 2025 Jonas Lin & Berry Feng). Feedback, issue reports, and community contributions are greatly appreciated to help expand the capabilities of GA-LLM.